\documentclass[sigconf,table]{acmart}

\usepackage[printonlyused,nolist,nohyperlinks]{acronym}

\AtBeginDocument{%
  }

\copyrightyear{2025}
\acmYear{2025}
\setcopyright{cc}
\setcctype{by}
\acmConference[FAccT '25]{The 2025 ACM Conference on Fairness, Accountability, and Transparency}{June 23--26, 2025}{Athens, Greece}
\acmBooktitle{The 2025 ACM Conference on Fairness, Accountability, and Transparency (FAccT '25), June 23--26, 2025, Athens, Greece}
\acmDOI{10.1145/3715275.3732087}
\acmISBN{979-8-4007-1482-5/2025/06}

\begin{document}

\title{Opening the Scope of Openness in AI}

\author{Tamara Paris}
\email{tamara.paris@mail.mcgill.ca}
\affiliation{%
  \institution{McGill University}
  \city{Montreal}
  \country{Canada}}

\author{AJung Moon}
\email{ajung.moon@mcgill.ca}
\authornote{Both authors contributed equally to this research.}
\affiliation{%
  \institution{McGill University}
  \city{Montreal}
  \country{Canada}
}

\author{Jin L.C. Guo}
\email{jguo@cs.mcgill.ca}
\authornotemark[1]
\affiliation{%
  \institution{McGill University}
  \city{Montreal}
  \country{Canada}
}

\begin{abstract}
The concept of \textit{openness} in AI has so far been heavily inspired by the definition and community practice of open source software. This positions openness in AI as having positive connotations; it introduces assumptions of certain advantages, such as collaborative innovation and transparency. However, the practices and benefits of open source software are not fully transferable to AI, which has its own challenges. Framing a notion of openness tailored to AI is crucial to addressing its growing societal implications, risks, and capabilities. We argue that considering the fundamental scope of openness in different disciplines will broaden discussions, introduce important perspectives, and reflect on what openness in AI should mean. Toward this goal, we qualitatively analyze 98 concepts of openness discovered from topic modeling, through which we develop a taxonomy of openness. Using this taxonomy as an instrument, we situate the current discussion on AI openness, identify gaps and highlight links with other disciplines. Our work contributes to the recent efforts in framing openness in AI by reflecting principles and practices of openness beyond open source software and calls for a more holistic view of openness in terms of actions, system properties, and ethical objectives.
\end{abstract}



\acrodef{AI}{Artificial Intelligence}
\acrodef{OSI}{Open Source Initiative}
\acrodef{OSS}{Open Source Software}
\acrodef{OSINT}{Open Source Intelligence}
\acrodef{IT}{Information Technology}
\definecolor{darkblue}{RGB}{27, 102, 215}
\definecolor{lightgrey}{RGB}{246, 246, 246}

\begin{teaserfigure}
    \centering
    \includegraphics[width=\textwidth]{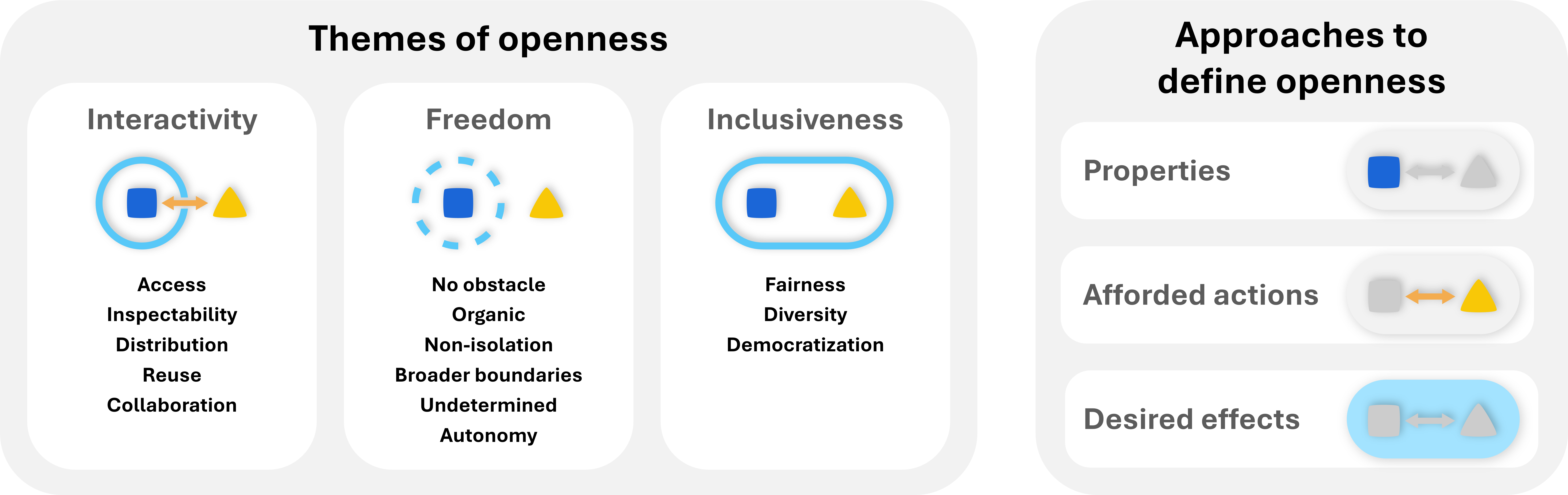}
    \caption{Overview of the taxonomy of openness, consisting of two dimensions: the themes of openness and the approaches to define openness. The themes include \textit{Interactivity} (Section \ref{sec:interactivity}), \textit{Freedom} (Section \ref{sec:freedom}) and \textit{Inclusiveness} (Section \ref{sec:inclusiveness}). The square represents \textit{what} is open and the triangle an \textit{external agent}.}
    \label{fig:themes}
    \Description{The two dimensions of the taxonomy of openness: the themes of openness and the approaches to define openness. On the left, the three main themes of the first dimension are illustrated through three pictograms, and the titles of the sub-themes are then listed. On the right, the three approaches of the second dimension are also illustrated with pictograms. Each pictogram is composed of a square and a triangle, with the square representing what is open and the triangle representing an external agent.}
\end{teaserfigure}

\maketitle

\section{Introduction}

Today, open source practices, such as making source code and components of an AI system publicly available, are often highly encouraged in the \ac{AI} community~\cite{Building_Open_Source_AI_2023,Zuckerberg_open_source_AI_path_forward_2024}. This is in part because of the crucial role open source practices play in advancing transparency, interoperability, and progress in \ac{AI}~\cite{using_proprietary_models_in_academic_2024,Building_Open_Source_AI_2023,Model_Openness_Framework_2024,Zuckerberg_open_source_AI_path_forward_2024}, similar to the benefits they offer for traditional software. However, existing  \ac{OSS} licenses and the open source definition proposed by \ac{OSI}~\cite{Perens_Sroka_Stu_1999} do not readily apply to \ac{AI} systems and their components --- AI systems are not only composed of source code but also of model weights, which are themselves generated by code and data, making them different from the usual structure of software~\cite{Defining_Open_Source_AI_news_2022,AI_weights_are_not_open_source_2023,Model_Openness_Framework_2024}. Moreover, the access to the code and weights without the documentation and training data is not sufficient to allow the freedoms that are supposed to be enabled by the open source practice (i.e., to study, modify, share and use the system). In 2022, the \ac{OSI} launched Deep Dive AI, a global multi-stakeholder co-design process to define open source AI~\cite{OSI_Open_Source_AI_def}, highlighting the need for a specific notion of openness to be tailored for the domain. The first version of the definition was then released in October 2024 and largely adapts the definition of \ac{OSS} to \ac{AI}; it requires code, parameters and data information to be made available in a manner that enables freedom to use for any purpose, study, modify, and share.

However, framing openness in \ac{AI} involves addressing multiple challenges beyond those of \ac{OSS}. First, because \ac{AI} systems and their development process are highly complex, the binary conception of openness (open vs. closed) as in the \ac{OSS} definition is too limiting to express the different types of components and parts of architectural elements that may be open~\cite{Solaiman_2023}. Hence, some suggest the adoption of a gradient and composite notion of openness for AI~\cite{Solaiman_2023,Openwashing_and_UE_Act_2024,towards_a_framework_for_Openness_2024}. Secondly, the benefits usually associated with \ac{OSS}, such as security, redistribution of influence, and transparency, are not completely transferable to \ac{AI}. For example, developing highly capable models requires significant resources that are only accessible to a limited number of actors; unlike \ac{OSS}, the availability of AI related code and knowledge does not mean that the capabilities of AI have been freely distributed or that the model can be reproduced~\cite{Open-Sourcing_Highly_Capable_Models_2023, open_for_business_2023}. Finally, given the documented risks associated with \ac{AI} models and their applications, the decision to make certain \ac{AI} open can involve a heightened degree of ethics considerations than those involved in traditional software~\cite{Balancing_transparency_and_risks_2025,Hazards_accessible_fine_tuning_2023,Open-Sourcing_Highly_Capable_Models_2023}. 
For example, misuse of open foundation models is one of the central risks of openness~\cite{Limits_and_possibilities_open_source_ddeepfakes_2022}, whereas this concern is not as prevalent for traditional software. 
Therefore, while the open source AI definition provided by the OSI is already a significant step forward for the \ac{AI} community, it should be complemented by considering other aspects of openness that extend both the practices and ethical objectives of \ac{OSS} to address these challenges specific to AI.

Looking beyond \ac{OSS}, many concepts related to openness in other domains can enrich the discussion of openness in \ac{AI}. For example, the Model Openness Framework introduced by~\citet{Model_Openness_Framework_2024} draws inspiration from concepts such as \textit{open science}, \textit{open access}, and \textit{open data} to suggest a more responsive openness framework for AI. Moreover, openness is a multifaceted concept that is not confined to computer science~\cite{openness_2022}. Its usage is grounded in a shared set of ideas, despite its specificity in each field. Previous work on openness in other disciplines has described some core aspects of openness, such as inclusivity~\cite{openness_and_the_intellectual_commons_2014}. Those aspects are also considered important for \ac{AI}~\cite{AI4ALL_2024}, but are not emphasized in the existing discussion concerning openness. Such an observation suggests that the current discourse of openness in AI is narrowly limited to certain aspects and misses significant interdisciplinary inputs to ensure its relevancy. 

In this study, we zoom out to take a broader and more interdisciplinary understanding of \textit{openness} to reflect on the current practice concerning openness in AI. We seek to inspire a discussion of what a truly open AI technology and community could and should mean. We first analyze openness concepts expressed across a wide range of academic literature, identified through topic modeling of a large multidisciplinary corpus of scientific articles. We then qualitatively analyzed 98 concepts that surfaced from this process and present it as a taxonomy of openness. The concepts include both abstract and concrete ones, allowing us to highlight properties  that are both physical and metaphorical. Finally, we situate \ac{AI} Openness within this broader taxonomy of openness, using which we highlight unexplored perspectives. 
Our work contributes to the responsible \ac{AI} community by providing a comprehensive taxonomy of openness that can be used by future work to examine practices and principles of openness in \ac{AI}. By highlighting the multiplicity of the nuances associated with this term, we call for a broader vision of openness beyond its current use in \ac{AI} that can promote discussions on tools and practices related to openness and their relation with ethical objectives. 

\section{Background}
\subsection{From OSS to Open Source AI}
At the dawn of computer commercialization in the 1950s, the source code of software was directly accessible to users. Developers collaborated with each other to create this code, marking the beginnings of open source development practices without being named as such~\cite{from_open_science_to_open_source_2021}. 
In the 1980s, in response to the rise of proprietary software, the free software movement emerged to promote these practices and advocated for everyone's freedom to run, modify, study, and share software code~\cite{Stallman_1985}. However, the philosophical and political stance of the free software movement made openness practices less appealing to the software industry. Thus, the term ``open source'' was created to distance itself from the connotations of free software while adopting its practices~\cite{history_of_OSI_2006}. Although the two movements share a common foundation (i.e., freedom of use, study, redistribute, and modify the source code), their goals diverge. Notably, free software focuses much more on protecting users' fundamental freedoms against corporate interests, whereas open source promotes a set of practices that can also be beneficial in pursuit of profit~\cite{open_source_misses_point_free_software_2009,Openness_with_and_without_IT_2017}.

The \ac{OSI} was created in 1998 to promote the open source philosophy, maintain a definition and approve associated licenses~\cite{Perens_Sroka_Stu_1999}.
Although open source as defined by the \ac{OSI} is a binary concept – either a piece of software is open source or it is not – there are still nuances of openness beyond this definition. For example, some open source licenses are copyleft, requiring that any software reusing the code, whether modified or not, can only be distributed if it preserves the same rights~\cite{what_is_copy_left}. Among proprietary software, some are referred to as ``source available'' when their code is visible to everyone, even though they do not provide the same freedoms guaranteed by open source~\cite{Source_Available_Licenses}. 

\ac{OSS} has been particularly beneficial in AI. For example, many open source machine learning development frameworks like PyTorch and TensorFlow helped accelerate technological advancement~\cite{How_open_source_shapes_AI_2022}. Numerous powerful \ac{AI} models have been introduced to the public as being ``open'' or ``open source'', often using these terms interchangeably~\cite{open_for_business_2023}. But, similar to how these terms have sometimes been misused to represent some software and data as ``open'' ~\cite{openwhasing_2017,fifty_shades_of_open_2016}, some organizations use them as a marketing ploy without meeting the expectation associated with these terms. This raised debates and criticisms about ``open-washing'' practices~\cite{Why_open_are_actually_close_2024,why_companies_democratise_AI_2024,Openwashing_and_UE_Act_2024,open_for_business_2023,Opening_up_chatGPT_2023}. 
For example, many AI models released as ``open'' restrict certain uses, do not release training source code, and/or do not make the necessary dataset publicly accessible~\cite{Opening_up_chatGPT_2023,Openwashing_and_UE_Act_2024} (as is the case with LLaMa~\cite{Llama2_license_2023,Llama_2_not_open_source_OSI_2023}).
At the same time, some ``open'' models do not strictly adhere to the usual practices associated with \ac{OSS} for different purposes. For example, Open Responsible AI Licenses (OpenRAIL) restrict certain uses of AI to prevent misuse while preserving many desirable properties of openness~\cite{openRAIL,RAIL_uses_2024,RAIL_intro_2022}.

Moreover, there is no clear consensus on what openness means in AI, making it more difficult to identify actual instances of open-washing. While the \ac{OSI} has contributed significantly to defining open source AI~\cite{OSI_Open_Source_AI_def}, it has faced numerous criticisms, particularly regarding the lack of requirements for data openness~\cite{OSI_readies_conversial_OSAID_2024,OSAID_is_not_fit_for_purpose_2024,ai_industry_is_trying_to_subvert_OSAID_2024}. The OSI’s definition is only one perspective among many others, exemplified by the multiple other frameworks that provide a way to assess openness with a less binary approach, such as the Model Openness Framework by~\citet{Model_Openness_Framework_2024}, the Community-Driven Assessment Framework by~\citet{Openwashing_and_UE_Act_2024}, and the classification scale proposed by~\citet{Risks_and_opportunities_2024}.

Beyond open-washing, ethical and security risks also poses critical concerns related to openness in AI~\cite{Balancing_transparency_and_risks_2025,on_the_societal_impact_2024,Risks_and_opportunities_2024,Open-Sourcing_Highly_Capable_Models_2023,Hazards_accessible_fine_tuning_2023,Near_to_mid_term_2024}.
For example, the emergence of harmful misuses of technology, such as non-consensual pornographic deepfakes, challenges the support for unbridled openness~\cite{Limits_and_possibilities_open_source_ddeepfakes_2022}. Balancing the risks and benefits of openness in AI is a crucial challenge for the responsible AI community; researchers are calling for greater caution in open publication of models with significant risks of misuse~\cite{on_the_societal_impact_2024,The_tension_between_openness_and_prudence_2020,offence_defense_balance_publishing_AI_2020,Open-Sourcing_Highly_Capable_Models_2023}. Many of these ethical concerns present challenges distinct from traditional software, such as greater capabilities or a higher concentration of resources required for development~\cite{open_for_business_2023}. Consequently, broadening the definition of openness beyond \ac{OSS} is not only necessary to critically consider \ac{AI}-specific technical realities, but also indispensable to mitigate the societal risks associated with these systems.

\subsection{Openness Across Disciplines}

From \textit{open education}, \textit{open science}, to \textit{open innovation}, the concept of openness can be found across disciplines. Even if openness is primarily contextual, prior studies have highlighted common principles among these concepts. For example, \textit{transparency}, \textit{access}, \textit{participation}, and \textit{democratization} are described in the openness framework presented by~\citet{Openness_with_and_without_IT_2017}, and \textit{inclusivity} and \textit{freedom} are highlighted by~\citet{openness_and_the_intellectual_commons_2014}. Moreover, even if these common principles and practices connect in various forms, the realization of these concepts does not always meet the ideals~\cite{openness_2022}. For example, although open source software development allows anyone to participate, making these practices theoretically ``inclusive,'' in practice women are often left out due to the barrier of sexism present in these communities~\cite{openness_2022,FLOSS_Inclusiveness_2019}. 

Openness, being extensively used throughout the literature, presents a real challenge in capturing all its nuances. Previous efforts have provided detailed understandings of many openness concepts and how they interact with one another~\cite{the_virtues_of_openness_2015,openness_2022}, while other have offered frameworks for this notion~\cite{Openness_with_and_without_IT_2017}. However, to the best of our knowledge, no existing work offers a thematic framework of openness that is both extensive and structured: previous work either focuses on a few aspects already known by the \ac{AI} community~\cite{Openness_with_and_without_IT_2017}, or it describes a wide range of concepts but lacks a precise structure that can be easily applied to the \ac{AI} domain~\cite{the_virtues_of_openness_2015}. 

\section{Method}

\begin{figure*}[t]
    \centering
    \includegraphics[width=\textwidth]{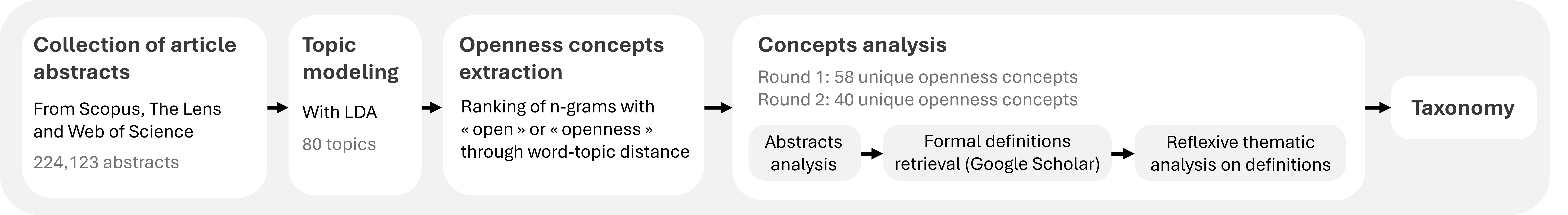}
    \caption{Summary of the method used to derive the taxonomy of openness.}
    \Description{Summary of the method section. The goal of this figure is to illustrate the 5 main steps of the method to derive the taxonomy of openness: collection of article abstracts, topic modeling, openness concepts extraction, concept analysis and taxonomy. The step of the concept analysis is sub-divided into 3 sub-steps: abstracts analysis, formal definition retrieval and reflexive thematic analysis on definition.}
    \label{fig:method}
\end{figure*}

In this study, we define \textit{openness} as encompassing all uses of the adjective ``open'' or the noun ``openness''. 
Consequently, an \textit{openness concept} refers to any unique construct in the academic literature containing these words. This includes both abstract concepts (e.g., \textit{open access} and \textit{trade openness}) and concrete concepts referring to tangible phenomena (e.g., \textit{open water} and \textit{open surgery}). 
The volume of literature related to openness is too extensive for a manual review (our final corpus contains 224,123 articles, Sect. \ref{sec:data_collection}). Hence, we employed topic modeling (Sect. \ref{sec:concept_extraction}) to discover a diverse set of \textit{openness concepts}, to distinguish between homonyms (e.g., \textit{open field} in agriculture versus in neurology), and to group closely related concepts (e.g., \textit{open circuit} and \textit{open phase fault}). 
Once we established a list of \textit{openness concepts} and retrieved their definitions, we conducted a reflexive thematic analysis on their definitions (Sect. \ref{sec:concept_definition}).
Figure \ref{fig:method} provides an overview of our method. 

\subsection{Collection of Article Abstracts}
\label{sec:data_collection}

We queried abstracts from three bibliographic databases: the Web of Science Core Collection, Scopus, and The Lens. These databases include open-access and subscription-based publications, representing a broad disciplinary coverage such as science, engineering, medicine, and the humanities~\cite{disciplinary_coverage_databases_2022,Google_scholar_to_overshadow_2019,comparaison_scopus_wos_2013,A_tale_two_databases_2020,Comparative_analysis_database_2024}.
Each database was queried with the term ``open|openness'' in the title. Our search was limited to journal, conference, or review articles written in English and published between 1980 and 2024. Bibliographic data of all matched entries were collected on September 5 and 6, 2024.  Entries without abstracts or with abstracts but shorter than 200 characters were removed. Following these steps, we obtained a total of $555,216$ article abstracts ($202,786$ from Scopus, $137,755$ from Web of Science, and $214,675$ from The Lens).
We then removed $267,734$ duplicates using unique identifiers, along with titles, author names, sources, and years of publication when no identifier was available. 
We also excluded $63,359$ articles that do not contain ``open|openness'' in the abstract, use ``open'' only as a verb in the title (e.g., ``Opening a door''), or appear as erroneous entries. 
Our resulting corpus contains $224,123$ bibliographic records.

\subsection{Extraction of the Openness Concepts}
\label{sec:concept_extraction}
We applied standard preprocessing steps on each record, including removing links, emails, HTML tags, punctuation, numbers, one- or two-character words, stopwords, credits, and disclosure sections. We also excluded keywords commonly found in abstracts that would introduce noise into our model, such as ``methodology'' and ``results''. We then performed lemmatization on the remaining text using Spacy~\cite{Spacy_2020}, and included 2-, 3- and 4-grams. Finally, any words and n-grams appearing less than 30 times were removed from the corpus.

We applied Latent Dirichlet Allocation (LDA) using Tomotopy~\cite{tomotopy_2022} to uncover the topics.
Through an iterative evaluation, we determined that 80 topics provided the best trade-off between separating concepts into distinct topics and reducing redundancy between topics. 
For each topic, we classified abstracts based on the document-topic distance and n-grams containing the terms ``open'' and ``openness'' using the word-topic distance. Among these n-grams, we identified at most two cohesive openness concepts per topic by reading abstracts associated with it. In total, 98 concepts were identified from the 80 topics (58 for the first round of analysis and 40 for the second). The complete list of topics and openness concepts is presented in Appendix \ref{sec:appendix}, along with other information related to the topic models.

\subsection{Qualitative Analysis of the Openness Concepts}
\label{sec:concept_definition}

Article abstracts rarely contain formal definitions. Therefore,
we used Google Scholar to retrieve formal definitions of each openness concept identified in our abstract-centric corpus. In particular, we queried each concept as a keyword alongside words that typically signal the occurrence of a definition, such as ``is defined,'' ``refers to,'' or preceded by ``the concept of'' or ``the term.'' This process yielded a collection of at least five formal definitions for each openness concept as expressed by the scholars of the relevant disciplines. We manually examined whether the definitions were consistent with the context of the abstracts previously read.
We then analyzed the resulting definitions through a reflexive thematic analysis process~\cite{using_thematic_analysis_2006,reflecting_on_thematic_analysis_2019}. We built a final taxonomy through two rounds of analysis and validation among all authors.

\section{Taxonomy of Openness}
\label{sec:taxonomy}

We organize our result as a taxonomy of openness along two dimensions: (1) the \textbf{\textit{themes}} covered by the openness concepts and (2) the \textbf{\textit{approaches}} used to define the openness concepts. The first dimension is composed of three main themes: \textit{Interactivity} (Section \ref{sec:interactivity}), \textit{Freedom} (Section \ref{sec:freedom}), and \textit{Inclusiveness} (Section \ref{sec:inclusiveness}), each themes encompassing several sub-themes.
The second dimension outlines the three approaches employed in the literature to define the openness concepts: through (1) \textit{properties} intrinsic to what is open (e.g.,``permeability'' and ``modular structure''), (2) \textit{afforded action} that openness enables or prevents (e.g.,``inspect'' and ``interfere''), and (3) \textit{desired effects} of the openness concepts (e.g., ``transparency'' and ``democratization''). Some concepts are only define through \textit{properties}, \textit{afforded actions}, or \textit{desired effects}, while others relate to multiple or all of these approaches. An overview of the two dimensions is provided in Fig.~\ref{fig:themes}.

Our taxonomy provides a detailed framework of how openness is defined in the literature, highlighting the thematic links between different aspects of openness and the cause-and-effect relationships. For example, the \textit{property} of ``modularity'' and the \textit{afforded action} of ``building add-on products'' found in the concept of \textit{open architecture} \cite{Open_architecture_1_2014,Open_architecture_2_2003} are not only examples of the theme ``reuse''; one is also an enabler for the other. When introducing our taxonomy, we also discuss how each theme and sub-themes are related to the existing discussion of openness in AI.

\subsection{Interactivity}
\label{sec:interactivity}

The openness concepts belonging to this theme are characterized by \textbf{interactions between inside and outside entities}, where ``inside'' refers to \textit{what} is made open. This includes interactions initiated from inside (e.g., distribution), outside (e.g., inspection), and both (e.g., collaboration). See Table \ref{tab:interactivity} for examples of \textit{properties}, \textit{afforded actions} and \textit{desired effects} for each sub-theme.

\begin{table*}
  \caption{Examples of properties, afforded actions and desired effects (\textcolor{darkblue}{blue}) extracted from definitions of openness concepts (\textit{italic}) illustrating each sub-theme of \textit{Interactivity} (Section \ref{sec:interactivity}).}
  \label{tab:interactivity}
  \begin{tabular}[x]{lll@{\hskip 0.08in}l@{\hskip 0.08in}ll}
    \toprule
    \rowcolor{white}
    \multicolumn{1}{l}{\textbf{Sub-theme}}&
    \multicolumn{1}{c}{\textbf{Properties}}&
    \multicolumn{3}{c}{\textbf{Afforded actions}}&
    \multicolumn{1}{c}{\textbf{Desired effects}}\\ \cmidrule(lr){3-5}

    \rowcolor{white}
    &
    &
    \multicolumn{1}{c}{\textbf{what}}&
    \multicolumn{1}{c}{\textbf{who}}&
    \multicolumn{1}{c}{\textbf{how}}&
    \\
    \midrule

    \rowcolor{lightgrey}
    Access& 
    \begin{tabular}[t]{@{}l@{}}
        Intelligence from\\
        publicly \textcolor{darkblue}{available} \\
        sources (\textit{OSINT})\\
        \textcolor{darkblue}{Incision} to\\
        access the organs\\
        (\textit{Open surgery})
    \end{tabular}& 
    \begin{tabular}[t]{l}
        \textcolor{darkblue}{Acquires}\\
        technology\\ 
        (\textit{Open innovation})\\
        \textcolor{darkblue}{Access} to the\\
        scholarly work\\
        (\textit{Open access})
    \end{tabular}& 
    &
    &
    \begin{tabular}[t]{@{}l@{}}
        \textcolor{darkblue}{Accessibility}\\
        of scientific\\
        knowledge\\
        (\textit{Open science})
    \end{tabular}\\[58pt]

    \rowcolor{white}
    Inspectability&
    \begin{tabular}[t]{@{}l@{}}
        \textcolor{darkblue}{Unconcealement}\\
        of the drug to\\
        the patient\\
        and investigator\\
        (\textit{Open label trial})
    \end{tabular}& 
    \begin{tabular}[t]{@{}l@{}}
        \textcolor{darkblue}{Inspect} the code\\
        (\textit{Open source})\\
        \textcolor{darkblue}{Replicate} the\\
        study\\
        (\textit{Open science})
    \end{tabular}& 
    &
    \begin{tabular}[t]{@{}l@{}}
        Discuss\\
        \textcolor{darkblue}{honestly}\\
        (\textit{Open dialogue})
    \end{tabular}& 
    \begin{tabular}[t]{@{}l@{}}
        More\\
        \textcolor{darkblue}{transparency}\\
        in research\\
        (\textit{Open science})\\
    \end{tabular}\\[47pt]

    \rowcolor{lightgrey}
    Distribution& 
    &
    \begin{tabular}[t]{@{}l@{}}
        Right to \textcolor{darkblue}{share}\\
        and \textcolor{darkblue}{redistribute}\\
        (\textit{Open source})
    \end{tabular}&
    & 
    & 
    \\[25pt]

    \rowcolor{white}
    Reuse& 
    \begin{tabular}[t]{@{}l@{}}
        \textcolor{darkblue}{Independent}\\
        \textcolor{darkblue}{modules} and\\
        \textcolor{darkblue}{standard} interfaces\\
        (\textit{Open architecture})
    \end{tabular}& 
    \begin{tabular}[t]{@{}l@{}}
        \textcolor{darkblue}{Integrate} in\\
        an other\\
        application\\
        (\textit{Open API})
    \end{tabular}& 
    & 
    &
    \begin{tabular}[t]{@{}l@{}}
        Maximizing\\
        \textcolor{darkblue}{interoperability}\\
        (\textit{Open data})
    \end{tabular}\\[36pt]

    \rowcolor{lightgrey}
    Collaboration& 
    & 
    \begin{tabular}[t]{@{}l@{}}
         \textcolor{darkblue}{Involve} external\\
         stakeholders\\ 
        (\textit{Open strategy})
    \end{tabular}& 
    \begin{tabular}[t]{@{}l@{}}
        The \textcolor{darkblue}{community}\\
        participate\\
        (\textit{Open source}\\
        \textit{development})
    \end{tabular}&
    \begin{tabular}[t]{@{}l@{}}
        Exchanges\\
        \textcolor{darkblue}{through}\\
        \textcolor{darkblue}{partnerships}\\ 
        (\textit{Open innovation})
    \end{tabular}&
    \\[36pt]
    
  \bottomrule
\end{tabular}
\end{table*}

\subsubsection{Access}
\label{sec:access}

Many openness concepts feature the ability or right of an external entity to access an internal element. The notion of access and availability are sometimes used interchangeably in the definitions of those concepts. In some contexts, access represents a stronger notion than availability: access can be afforded by a practicable entry (e.g., an incision to access the organs in \textit{open surgery}~\cite{Open_surgery_1_2019,Open_surgery_2_2022}) or an identifiable access-path (e.g., a link to access the full text in \textit{open access}~\cite{BOAI,Open_access_2_2020}), which is not guaranteed by availability. For example, \textit{open source intelligence} (OSINT) is defined as intelligence ``produced from publicly available sources''~\cite{Böhm_Lolagar_2021}. This definition encompasses sources that cannot be easily found through major search engines~\cite{Balaji_Karmel_2022}. OSINT is therefore a concept that emphasizes availability rather than accessibility, unlike \ac{OSS}, where the means to access the source code must be well publicized~\cite{Perens_Sroka_Stu_1999}. Accessibility can also describe the \textit{desired effect} of the openness concept (e.g., \textit{open science}) rather than an \textit{afforded action}, then referring to the ease of access rather than access itself (e.g., \textit{open science}~\cite{Open_science_1_2023}).

\textbf{\textit{Access} in \ac{AI} Openness:} Which components of an AI system are accessible and how they are accessible are key considerations in AI governance~\cite{position_paper_model_access_2024}. Prior studies that primarily focused on assessing the risks of openness typically include in their scope all models with accessible model weights and architecture~\cite{on_the_societal_impact_2024,Open-Sourcing_Highly_Capable_Models_2023}, while efforts to define AI openness such as the \ac{OSI}'s Open Source AI Definition~\cite{OSI_Open_Source_AI_def} and the Model Openness Framework~\cite{Model_Openness_Framework_2024} include access to additional components in their scope, such as training code, dataset descriptions, evaluation results, etc. Moreover, the release options for \ac{AI} models are varied~\cite{Solaiman_2023}, so access itself can refer to different practices, such as API call or downloading components. However, not all types of access provide the same guarantees. For example, being able to download the model avoids relying on third-party tools, provides autonomy and privacy, which is not guaranteed when using APIs~\cite{on_the_societal_impact_2024,Zuckerberg_open_source_AI_path_forward_2024}. Other types of access, such as through indexed and version controlled public code repositories are possible, but many models claimed ``open'' do not implement these types of access~\cite{Openwashing_and_UE_Act_2024,Opening_up_chatGPT_2023}.

\subsubsection{Inspectability}
\label{sec:inspect}

Openness sometimes refers to the possibility or the right of an external entity to inspect what is inside. This can be achieved by several \textit{properties} such as non-concealment (e.g., \textit{open label}~\cite{Open_label_1_2011,Open_label_2_2016}), unobstructed access (e.g., \textit{open heart surgery}~\cite{Open_heart_surgery_2014}), honesty (e.g., \textit{open dialogue}~\cite{Open_dialogue_1_2024}), non-secrecy (e.g., \textit{open strategy}~\cite{Open_strategy_1_2019}), and can be facilitated by detailed documentation (e.g., \textit{open source hardware}~\cite{open_source_hardware_3}). Several openness concepts are defined by the \textit{desired effects} enabled by inspection, such as transparency (e.g., \textit{open science}~\cite{Open_science_1_2023}, \textit{open government}~\cite{Open_government_1_2024}, \textit{open strategy}~\cite{Open_strategy_1_2019}), reproducibility and reliability (e.g., \textit{open science}~\cite{Open_science_1_2023}), or accountability (e.g., \textit{open government}~\cite{Open_Government_2_2019}). 

\textbf{\textit{Inspectability} in \ac{AI} Openness:} Inspection is often phrased as one benefit of allowing access to AI systems and a fundamental step for AI auditability, reproducibility, trust, and verifiability of the system~\cite{Why_open_are_actually_close_2024,using_proprietary_models_in_academic_2024,on_the_societal_impact_2024,Risks_and_opportunities_2024}, along with reducing the risk of major security, safety, or privacy flaws~\cite{Balancing_transparency_and_risks_2025,considerations_for_governing_open_2024,Hazards_accessible_fine_tuning_2023}. While inspection can be facilitated by making model cards~\cite{modelcard_2019}, datasheets~\cite{datasheets_2021}, and data cards~\cite{datacard_2022} available to the public -- which are sometimes included as requirements in the definitions of open models~\cite{Model_Openness_Framework_2024,OSI_Open_Source_AI_def} -- this documentation does not always translate to auditability and reproducibility of the model, that can be limited by resources~\cite{open_for_business_2023}. Moreover, the quality of the documentation and the trend of disseminating information via blog posts or preprints rather than peer-reviewed articles is a big concern for transparency~\cite{Openwashing_and_UE_Act_2024}.

\subsubsection{Distribution}
\label{sec:exchange}

Openness can be characterized by the cross-boundary exchange of resources, services (e.g., \textit{open market}~\cite{trade_openness_1_2020, trade_openness_2_2016}), and knowledge (e.g., \textit{open innovation}~\cite{Open_innovation_1_2011,Open_innovation_2_2018}). It includes the right and the possibility to share, redistribute (e.g., \textit{open educational resources}~\cite{open_educational_ressources_1,open_educational_ressources_4}), or sell even without modifications (e.g., \textit{open source hardware}~\cite{open_source_hardware_3,open_source_hardware_7}). These exchanges can be commercial (e.g., \textit{open innovation}~\cite{Open_innovation_1_2011,Open_innovation_2_2018}) or purely altruistic (e.g., sharing ideas in \textit{open education}~\cite{open_education_1,open_education_2}). Some forms of redistribution require acknowledgment of the source (e.g., \textit{open access}~\cite{Open_access_1_2015,Open_access_2_2020}) or the right to maintain the same license (e.g., \textit{\ac{OSS}}~\cite{Perens_Sroka_Stu_1999}). These exchanges are often factors that then enable reuse and collaboration. 

\textbf{\textit{Distribution} in \ac{AI} Openness:} The possibility and right to share is an integral part of the definition of \textit{open source \ac{AI}} as defined by the \ac{OSI}~\cite{OSI_Open_Source_AI_def}, and the right to redistribution is also mentioned in the Model Openness Framework~\cite{Model_Openness_Framework_2024}. However, these two frameworks do not specify whether the model provenance must be acknowledged.

\subsubsection{Reuse}
\label{sec:reuse}

Reuse includes the external use of internal resources or the internal use of external resources, with or without modification (e.g., \textit{open source hardware}~\cite{open_source_hardware_3,open_source_hardware_7} and \textit{open educational resources}~\cite{open_educational_ressources_1,open_educational_ressources_4}). Reuse can be facilitated by \textit{properties} such as modularity, standardization, and consensual practices (e.g., \textit{open architecture}~\cite{Open_architecture_1_2014,Open_architecture_2_2003}). For example, some services are designed to be integrated into another project (e.g., \textit{open APIs}~\cite{open_API_2,open_API_1_2018}) or to serve as portable solutions (e.g., \textit{open standards}~\cite{open_standard_1_2008}). Through interoperability, these practices permit to build new works on top of existing works (e.g., \textit{open data}~\cite{open_data_1_2022}).

\textbf{\textit{Reuse} in \ac{AI} Openness:} Reuse and customization of open models is a widely covered topic in the literature on \ac{AI} openness, and presented as an advantage for accelerating innovation, research, and the adoption of state-of-the-art models~\cite{Balancing_transparency_and_risks_2025,Model_Openness_Framework_2024,Risks_and_opportunities_2024,on_the_societal_impact_2024,why_companies_democratise_AI_2024,Open-Sourcing_Highly_Capable_Models_2023,considerations_for_governing_open_2024,Building_Open_Source_AI_2023}. 
Widespread use of a model would also make it possible to test the model limits~\cite{Why_open_are_actually_close_2024}, and improve its quality~\cite{why_companies_democratise_AI_2024}. At the same time, the reuse of AI models -- particularly foundation models -- carries serious risks such as harmful misuse, environmental impacts, and fairness issues~\cite{on_the_opportunities_risks_fundation_models_2022,Sociotechnical_harms_taxonomy_2023}. Openness can facilitate these misuse, for example by allowing ethical guidelines bypass~\cite{Risks_and_opportunities_2024} or by fine-tuning and combining open models with other tools to introduce new vulnerabilities or uses not intended by the developers~\cite{Open-Sourcing_Highly_Capable_Models_2023}, questioning its responsibility in case of harms~\cite{considerations_for_governing_open_2024}. Beyond the use of an open model's weights for downstream tasks, reusing other components becomes more challenging with limited resources~\cite{open_for_business_2023}.

\subsubsection{Collaboration}
\label{sec:collaborate}

Many openness concepts encompass participatory and collaborative practices. For example, external stakeholders can contribute to decisions and tasks that are usually reserved to people inside the organization (e.g. \textit{open strategy}~\cite{Open_strategy_1_2019}). It can also refer to collaboration, cooperation, and partnerships between entities, as opposed to them operating independently (e.g. \textit{open business model}~\cite{open_business_model_1_2014}). This enables entities to form networks and communities whith peer-recognition (e.g. \textit{open source software development}~\cite{open_source_software_dev_1_2018,open_source_software_dev_2_2011}).

\textbf{\textit{Collaboration} in \ac{AI} Openness:} Collaboration is often presented as a benefit of \ac{AI} Openness rather than as a part of its definition. 
For example, the \ac{OSI} definition~\cite{OSI_Open_Source_AI_def} mentions the benefit of ``collaborative improvement'' enabled by Open Source \ac{AI}, but does not require an \ac{AI} system to be built or maintained collaboratively to be labeled as ``open source.'' Similarly, the reuse of external machine learning models allows developers and researchers to collaborate by making practical advancement and scientific findings on top of existing ones~\cite{Open-Sourcing_Highly_Capable_Models_2023,Model_Openness_Framework_2024}. However, these forms of collaboration remain limited: they rely on a closed entity that retains control over any modifications made to the original model. Additionally, even if Hugging Face is often used to share \ac{AI} models and collaborate, the most popular models on this platform are often developed by large companies with little external collaboration~\cite{AI_community_building_future_2024}.

In parallel, some works emphasize the principle of open collaboration in \ac{AI} Openness, which allows a wider and more diverse audience to freely collaborate to create and maintain \ac{AI} systems by lowering the barriers to participate in project developments~\cite{Open_AI_Collaboratives_2023,BigScience_Case_Study_2022}. For example, the BigScience Workshop, an open research collaboration initiative, demonstrated the effectiveness of such practices in building large datasets~\cite{ROOTS_2023} and models~\cite{BLOOM_2023}. 

\subsection{Freedom}
\label{sec:freedom}
The openness concepts belonging to this theme are characterized by the \textbf{reduction of interference, requirements, obstacles, governance, and constraining structures}, allowing for more freedom and possibilities, as well as more organic evolutions. See Table~\ref{tab:freedom} for examples of \textit{properties}, \textit{afforded actions} and \textit{desired effects} for each sub-theme of \textit{freedom}.

\begin{table*}
  \caption{Examples of principles and practices (\textcolor{darkblue}{blue}) extracted from definitions of openness concepts (\textit{italic}) illustrating each sub-theme of \textit{Freedom} (Section \ref{sec:freedom})}
  \label{tab:freedom}
  \begin{tabular}[x]{lll@{\hskip 0.08in}l@{\hskip 0.08in}ll}
    \toprule
    \rowcolor{white}
    \multicolumn{1}{l}{\textbf{Sub-theme}}&
    \multicolumn{1}{c}{\textbf{Properties}}&
    \multicolumn{3}{c}{\textbf{Afforded actions}}&
    \multicolumn{1}{c}{\textbf{Desired}}\\ \cmidrule(lr){3-5}

    \rowcolor{white}
    &
    &
    \multicolumn{1}{c}{\textbf{what}}&
    \multicolumn{1}{c}{\textbf{who}}&
    \multicolumn{1}{c}{\textbf{how}}&
    \multicolumn{1}{c}{\textbf{effects}}\\
    \midrule

    \rowcolor{lightgrey}
    No obstacle& 
    \begin{tabular}[t]{@{}l@{}}
        \textcolor{darkblue}{Empty} and\\
        \textcolor{darkblue}{featureless} space\\
        (\textit{Open field test})
    \end{tabular}&
    & 
    &
    \begin{tabular}[t]{@{}l@{}}
        Access \textcolor{darkblue}{without}\\
        \textcolor{darkblue}{any charges}\\
        (\textit{Open access})
    \end{tabular}&
    \\[25pt]

    \rowcolor{white}
    Organic&
    \begin{tabular}[t]{@{}l@{}}
        \textcolor{darkblue}{Reduction} of\\
        \textcolor{darkblue}{regulation} and \textcolor{darkblue}{policy}\\
        (\textit{Open market})
    \end{tabular}&
    \begin{tabular}[t]{@{}l@{}}
        Regulation body\\
        \textcolor{darkblue}{do not interfer}\\
        (\textit{Open market})
    \end{tabular}&
    &
    \begin{tabular}[t]{@{}l@{}}
        Can be \textcolor{darkblue}{freely} used\\
        by all citizens\\
        (\textit{Urban open space})
    \end{tabular}&
    \begin{tabular}[t]{@{}l@{}}
        Provide\\
        \textcolor{darkblue}{learning}\\
        \textcolor{darkblue}{freedom}\\
        (\textit{MOOC})
    \end{tabular}\\[36pt]

    \rowcolor{lightgrey}
    Non-isolation& 
    \begin{tabular}[t]{@{}l@{}}
        \textcolor{darkblue}{Non-solid boundary}\\
        (\textit{Open boundary})\\
        \textcolor{darkblue}{Disconnected} circuit\\
        (\textit{Open circuit})
    \end{tabular}& 
    \begin{tabular}[t]{@{}l@{}}
        Energy \textcolor{darkblue}{leaks}\\
        (\textit{Open resonator})\\
        Air \textcolor{darkblue}{flows} in/out\\
        (\textit{Open cathode})
    \end{tabular}&
    & 
    &
    \\[36pt]

    \rowcolor{white}
    \begin{tabular}[t]{@{}l@{}}
        Broader\\
        boundaries\\
    \end{tabular}& 
    \begin{tabular}[t]{@{}l@{}}
        \textcolor{darkblue}{Broad-minded}\\
        (\textit{Openness to}\\
        \textit{experience})\\
    \end{tabular}&
    &
    &
    \begin{tabular}[t]{@{}l@{}}
        Can be solved\\
        \textcolor{darkblue}{in multiple ways}\\
        (\textit{Open-ended}\\
        \textit{problem})
    \end{tabular}&
    \begin{tabular}[t]{@{}l@{}}
        Promote\\
        students\\
        \textcolor{darkblue}{creativity}\\
        (\textit{Open education})
    \end{tabular}\\[36pt]

    \rowcolor{lightgrey}
    Undetermined& 
    \begin{tabular}[t]{@{}l@{}}
        \textcolor{darkblue}{No expectation}\\
        (\textit{Open-ended})\\
        \textcolor{darkblue}{Unsolved} problem\\
        (\textit{Open problem})
    \end{tabular}& 
    \begin{tabular}[t]{@{}l@{}}
        \textcolor{darkblue}{Adapt} to data\\
        it was not train\\
        to classify\\
        (\textit{Open set (ML)})
    \end{tabular}& 
    &
    &
    \\[36pt]

    \rowcolor{white}
    Autonomy& 
    \begin{tabular}[t]{@{}l@{}}
        \textcolor{darkblue}{Independent} thinker\\
        (\textit{Openness to}\\
        {experience})
    \end{tabular}& 
    \begin{tabular}[t]{@{}l@{}}
        Students \textcolor{darkblue}{choose}\\
        their approach\\
        (\textit{Open-ended}\\
        \textit{problem})
    \end{tabular}&
    \begin{tabular}[t]{@{}l@{}}
        Students manage\\
        their education\\
        \textcolor{darkblue}{themselves}\\
        (\textit{Open education})
    \end{tabular}& 
    &
    \begin{tabular}[t]{@{}l@{}}
        Promote\\
        students\\
        \textcolor{darkblue}{autonomy}\\
        (\textit{Open education})\\
    \end{tabular}\\[36pt]
    
  \bottomrule
\end{tabular}
\end{table*}

\subsubsection{No obstacle}
\label{sec:noobstacle}

Openness can refer to the removal of obstacles (e.g., obstacles blocking access to scientific information in \textit{open access}~\cite{Open_access_2016}), a low density of obstacles (e.g., \textit{open habitat}~\cite{open_habitat_1_2010,open_habitat_2_2024}), or even emptiness (e.g., \textit{open field test}~\cite{open_field_test_1_2019,open_field_test_2_2013}). For instance, \textit{open university} remove formal entry requirements and other technical barriers such as time, pace, and place constraints~\cite{open_univeristy_1,open_university_2_2016}. 
The obstacles removed can also include abstract barriers (e.g., judgment in \textit{open dialogue}~\cite{open_dialogue_2}) or tangible obstacles (e.g., ice in an \textit{open ocean}~\cite{open_ocean_ice_1_2024,open_ocean_ice_2_2021}). 

\textbf{\textit{No obstacle} in \ac{AI} Openness:} The main obstacle removed in \ac{AI} openness is the cost; most models considered ``open'' allow users to download their weights for free~\cite{Openwashing_and_UE_Act_2024}. However, price is not the only limitation in practice: financial and computational resources, talent, and high-quality data resources are concentrated in few academics or industry AI research labs~\cite{Open-Sourcing_Highly_Capable_Models_2023}, and the barriers to joining these labs is a major obstacle to participate in innovation~\cite{Open_AI_Collaboratives_2023}. Training a large language model requires considerable hardware resources, which can be difficult to access outside those labs~\cite{Laboratory_Scale_AI_2024,Why_open_are_actually_close_2024}; even with the emergence of decentralized methods for training models, such as cloud sourcing, might lower this barrier~\cite{Hazards_accessible_fine_tuning_2023}. 

\subsubsection{Organic}
\label{sec:organic}

Freedom in openness can be exemplified by the removal of external interferences and forms of governance to allow an organic evolution. For example, individuals are free to move within the natural space without artificial blockage or prohibition (e.g., \textit{urban open spaces}~\cite{urban_open_space_1,urban_open_space_2}). Unlike \textit{no obstacle} (Section \ref{sec:noobstacle}), this sub-theme does not necessarily imply the absence of obstacles. For instance, the removal of tariffs, subsidies, and quotas in an \textit{open market} allows the market to evolve organically by enabling natural competition to take place~\cite{open_market_2022}. However, obstacles might still form organically, potentially hindering the participation of certain actors. 
Organic freedom can also exist within regulations that are neutral and less interventionist (e.g., \textit{open source licenses}~\cite{Perens_Sroka_Stu_1999} do not restrict types of use). 

\textbf{\textit{Organic} in \ac{AI} Openness:} \ac{OSS}, which greatly inspires openness in \ac{AI}, contains several organic elements: it requires software to be neutral regarding the reuse; it places no constraints on individuals, technologies, or domains that will reuse its source code~\cite{Perens_Sroka_Stu_1999}. Open source developers assume that it is not their role to dictate how their code is used. However, many \ac{AI} models labeled as ``open,'' such as LLaMa 2, restrict specific types of use in their license~\cite{Llama_2_not_open_source_OSI_2023,Llama2_license_2023}. Many other open models opt for Responsible AI Licenses (RAIL), which is also restrictive in its allowed use cases~\cite{RAIL_intro_2022,RAIL_uses_2024}. This lack of neutrality, combined with the lack of accessible components and information in some open models, has sparked debates about the loose use of ``open'' and ``open source,'' qualifying them as ``open-washing''~\cite{Openwashing_and_UE_Act_2024}.

At the same time, the misuse of certain open foundation models is a genuine concern~\cite{Open-Sourcing_Highly_Capable_Models_2023,on_the_societal_impact_2024}, which drives some developers to restrict their usage. These restrictions are often implemented through licenses. The emergent efforts, such as self-destructing models~\cite{self_destructing_models_2023}, demonstrate an attempt to establish technical barriers as well. Moreover, some researchers advocate for greater caution in scientific publications on AI~\cite{The_tension_between_openness_and_prudence_2020,offence_defense_balance_publishing_AI_2020}. The irreversibility of openness makes it impossible to moderate harmful uses of models once released openly~\cite{Risks_and_opportunities_2024,Open-Sourcing_Highly_Capable_Models_2023,on_the_societal_impact_2024}, and can discourage some developers from making their models publicly accessible. The risk of misuses is not unique to AI: similar debates have already taken place about offensive cybersecurity tools that are open source~\cite{open_source_cybersecurity_deepdive_2023}. However, some AI models can be used in a wider range of situations than software can, making this risk more prominent in AI.

\subsubsection{Non-isolation} 
\label{sec:nonisolation}

Openness sometimes refers to a system that is not entirely isolated, characterized by the absence of a solid boundary (e.g., \textit{open boundary}~\cite{open_boundaries_1,open_boundaries_2}), permeability (e.g. \textit{open framework}~\cite{open_framework_2003}), porosity (e.g. \textit{open cell foam}~\cite{open_cell_foam_1,open_cell_foam_2}), or disconnections (e.g., \textit{open circuit}~\cite{open_circuit_1,open_circuit_2}). The state of non-isolation can provide passage between the outside and the inside, including uncontrolled passage. Such systems are prone to leakage (e.g., \textit{open resonator}~\cite{open_resonator_1,open_resonator_2}), exchange with the external environment (e.g., \textit{open fracture}~\cite{open_fracture_1,open_fracture_2}), or discharge of internal resources outside the system (e.g., \textit{open cycle}~\cite{open_cycle_2023}). Non-isolation can also result in the interruption of internal flows, as exemplified by the disruption of electrical current in an \textit{open circuit}~\cite{open_circuit_1,open_circuit_2}. This concept of openness sometimes connotes vulnerability and risky exposure, as seen in cases such as \textit{open fractures}~\cite{open_fracture_1,open_fracture_2}.

The sub-themes under \textit{interactivity} (Section \ref{sec:interactivity}), such as \textit{access} (Section \ref{sec:access}) or \textit{inspect} (Section \ref{sec:inspect}), can lead to non-isolation. What differentiates non-isolation from interactivity is its involuntary nature. At the same time, voluntary interactions can cause similar effects of non-isolation, such as unintended leakages. For example, publicly accessible data, such as open data, are vulnerable to personal information leaks by cross joining different datasets~\cite{Kao_Hsieh_Chu_Kuang_Yang_2017}.
In general, allowing external entities to inspect an internal system exposes it to security threats and potential leaks.

\textbf{\textit{Non-isolation} in \ac{AI} Openness:} Open models present risks of component leakage and security flaws due to their permeability and exposure~\cite{Balancing_transparency_and_risks_2025}. For example, white-box attacks are attacks on \ac{AI} models that are only possible if the inner works of the system, such as its architecture, are known to the attackers~\cite{adversial_ml_taxonomy_2024,hotflip_white_box_adversial_examples_2018}. Even if only the weights of a system are open, data and model components might still be recovered through the use of the model~\cite{stealing_ml_models_via_APIs_2016,stealing_part_language_model_2024,grey_box_extraction_2021}. Open collaboration in \ac{AI} can also allow some permeability and facilitate poisoning attacks~\cite{Balancing_transparency_and_risks_2025}. On the other hand, non-isolation could act as a cause of openness: it is not uncommon to see proprietary models leak to the public~\cite{Llama_leak_2023,OpenAI_leak_2024}, and the large number of people involved in the creation of AI models makes these leaks more difficult to prevent.

\subsubsection{Broader boundary}
\label{sec:broaderboundary}

The broadening of edges, limits, or boundaries is a recurring theme in openness. This refers to spatial expansion (e.g., \textit{open cluster}~\cite{open_cluster_1,open_cluster_2}) or the broadening of possibilities by reducing constraints (e.g., \textit{open-ended problem} have no constraints on their solution approach~\cite{open_ended_1,open_ended_2}). In this sub-theme, openness can be defined as the opposite of a narrower, closed state (e.g., \textit{open-angle glaucoma}~\cite{Open_angle_glaucoma_2022}). This broadening also refers to abstract notions, such as the enhancement of creativity and imagination (e.g. \textit{openness to experience}~\cite{openness_to_experience_1,openness_to_experience_2}). 

\textbf{\textit{Broader boundary} in \ac{AI} Openness:} In prior sections, we include within AI openness previous AI articles that use the terms ``open source,'' ``open,'' or ``openness.'' However, the term ``open-ended'' is also found in the literature, describing a machine learning model capable of generating sequences that are both novel and learnable by an observer~\cite{open_endedness_is_essential_2024,open_ended_AI_2}. We have chosen to separate this concept from the rest of AI openness because its definition is distinct from other openness-related terms in AI. However, open-ended models encapsulate the idea of pushing boundaries beyond what is already known and learned, making it a relevant concept for the sub-theme of \textit{broader boundaries}. The intersection between open-ended models and AI openness could play a big role in future research on risk of open foundation models, especially when considering concerns about the harmful misuse enabled by access.

\subsubsection{Undetermined}
\label{sec:undermined}

Openness can refer to a currently undetermined event that could be determined in the future (e.g., \textit{open problem}) or a space with undefined, indeterminate, or indeterminable boundaries (e.g., \textit{open set} in machine learning~\cite{open_set_ML_1,open_set_ML_2}). It includes areas outside the boundaries that can not be precisely delineate, such as \textit{urban open space}, which includes all urban areas outside of buildings~\cite{urban_open_space_1,urban_open_space_2}. Undetermined also refers to the absence of expectations regarding a future event (e.g., \textit{open-ended games} in education~\cite{open_ended_games_2020}). This leads to adaptability to future events that are impossible to predict, exemplified by \textit{open set} in machine learning, which refers to models encountering examples they were not trained to classify before~\cite{open_set_ML_1,open_set_ML_2}.

\textbf{\textit{Undetermined} in \ac{AI} Openness:} Similar to what was previously described in the sub-theme \textit{broader boundaries} (Section \ref{sec:broaderboundary}), the concept of open-ended models in AI closely aligns with \textit{undetermined}. Open-ended models produce novel outcomes, which therefore could not have been determined by the observer beforehand~\cite{open_endedness_is_essential_2024}. However, the inability to predict or determine the open-ended models' outcome, as well as the irreversibility of this process, also greatly amplifies the risks associated with those models.

\subsubsection{Autonomy}
\label{sec:autonomy}

In openness, autonomy refers to the freedom to act according to its own rules, without the need of permission. Student's autonomy is one of the core principles of \textit{open education}~\cite{open_education_1,open_education_2}. Students are able to make their own choices, without being constrained to a particular pace of work or a prescribed curriculum. Students are responsible for their own education without a higher authority dictating decisions. Similarly, autonomy presents in many other education-related concepts, such as \textit{open university}~\cite{open_univeristy_1} and \textit{open-ended games}~\cite{open_ended_games_2020}.
Autonomy is also influenced by other principles of openness, such as inspectability (Section~\ref{sec:inspect}), which allows individuals to be well-informed to make their own decisions. Autonomy also promotes individual empowerment: for example, \textit{open banking} gives customers the power to decide whether to allow third parties to access their data~\cite{open_banking_1,open_banking_2}.

\textbf{\textit{Autonomy} in \ac{AI} Openness:} 
\ac{OSI}'s Open Source AI Definitions explicitly state that components can be reused without asking for permission~\cite{OSI_Open_Source_AI_def}. The ability for developers to download AI components and make their own copies increases developers' autonomy. For example, developers can choose not to update these components~\cite{Risks_and_opportunities_2024}. 

\subsection{Inclusiveness}
\label{sec:inclusiveness}
The openness concepts belonging to this theme are characterized by the \textbf{inclusion of each individual in a fair and responsive manner}. This includes representativeness, eliminating discrimination, seeking diversity, recognizing individual needs, and democratization. Inclusiveness can also apply to non-individual entity such as countries, companies or scientific fields. See Table~\ref{tab:inclusiveness} for examples of \textit{properties}, \textit{afforded actions} and \textit{desired effects} for each sub-theme.

\begin{table*}
  \caption{Examples of principles and practices (\textcolor{darkblue}{blue}) extracted from definitions of openness concepts (\textit{italic}) illustrating each sub-theme of \textit{Inclusiveness} (Section \ref{sec:inclusiveness})}
  \label{tab:inclusiveness}
  \begin{tabular}[x]{lll@{\hskip 0.08in}l@{\hskip 0.08in}ll}
    \toprule
    \rowcolor{white}
    \multicolumn{1}{l}{\textbf{Sub-theme}}&
    \multicolumn{1}{c}{\textbf{Properties}}&
    \multicolumn{3}{c}{\textbf{Afforded actions}}&
    \multicolumn{1}{c}{\textbf{Desired effect}}\\ \cmidrule(lr){3-5}

    \rowcolor{white}
    &
    &
    \multicolumn{1}{c}{\textbf{what}}&
    \multicolumn{1}{c}{\textbf{who}}&
    \multicolumn{1}{c}{\textbf{how}}& \\
    \midrule

    \rowcolor{lightgrey}
    Fairness& 
    \begin{tabular}[t]{@{}l@{}}
        Licensed on \textcolor{darkblue}{non-}\\
        \textcolor{darkblue}{discriminatory} terms\\
        (\textit{Open standard})\\
        \textcolor{darkblue}{Unbiased} attitude\\
        (\textit{Open minded})
    \end{tabular}& 
    & 
    & 
    \begin{tabular}[t]{@{}l@{}}
        Enter in the\\
        market \textcolor{darkblue}{with}\\
        \textcolor{darkblue}{equal}\\
        \textcolor{darkblue}{opportunity}\\
        (\textit{Open market})
    \end{tabular}& 
    \begin{tabular}[t]{@{}l@{}}
        Strives to meet\\ 
        the ideal of\\
        \textcolor{darkblue}{objectivity}\\
        and \textcolor{darkblue}{impartiality}\\
        (\textit{Open minded})
    \end{tabular}\\[47pt]

    \rowcolor{white}
    Diversity&
    \begin{tabular}[t]{@{}l@{}}
         Curriculum \textcolor{darkblue}{variety}\\
        (\textit{Open univeristy})\\
        Preference for \textcolor{darkblue}{variety}\\
        (\textit{Openness to experience})
    \end{tabular}&
    &
    \begin{tabular}[t]{@{}l@{}}
        \textcolor{darkblue}{Diverse}\\
        \textcolor{darkblue}{stakeholders}\\
        participate\\
        (\textit{Open strategy})
    \end{tabular}&
    &
    \\[36pt]

    \rowcolor{lightgrey}
    Democratization& 
    \begin{tabular}[t]{@{}l@{}}
        \textcolor{darkblue}{De-centered}, \textcolor{darkblue}{distributed}\\
        and \textcolor{darkblue}{non-hierarchical}\\
        development process\\
        (\textit{Open source development})
    \end{tabular}& 
    \begin{tabular}[t]{@{}l@{}}
        \textcolor{darkblue}{Diffuse}\\
        scientific\\
        knowledge\\
        (\textit{Open science})
    \end{tabular}&  
    \begin{tabular}[t]{@{}l@{}}
        \textcolor{darkblue}{Anyone} can\\
        access to the\\
        academic work\\
        (\textit{Open access})
    \end{tabular}& 
    &
    \begin{tabular}[t]{@{}l@{}}
        \textcolor{darkblue}{Democratization}\\
        of education\\
        (\textit{Open education})
    \end{tabular}\\[36pt]
    
  \bottomrule
\end{tabular}
\end{table*}

\subsubsection{Fairness}
\label{sec:fairness}

Equality, equity, and justice are common principles across several concepts of openness. For example, access to \textit{open source software}~\cite{Perens_Sroka_Stu_1999}, \textit{open educational resources}, and ~\cite{open_educational_ressources_1} \textit{open standards} should be non-discriminatory towards groups of individuals, ensuring equality for all. However, equality does not always imply equity or justice. For example, \textit{open market} promotes equal opportunities but does not involve equity~\cite{open_market_4}. In contrast, \textit{open education} promotes equity and justice among students, acknowledging that some individuals have greater needs than others~\cite{open_education_5}. Fairness also encompasses impartiality and objectivity, exemplified by \textit{open-mindedness}, which refers to an individual's willingness to consider opposing opinions without biases~\cite{open_mindedness_1,open_mindedness_2}.

\textbf{\textit{Fairness} in \ac{AI} Openness:} To some extend, openness can help mitigating inequality between social groups; for example, inspectability enabled by openness of \ac{AI} models allows detecting biases or harms toward marginalized groups, thereby improving the fairness of the models~\cite{Model_Openness_Framework_2024,on_the_societal_impact_2024}. Openness in \ac{AI} also facilitates the redistribution of access to state-of-the-art technologies, helping to close the inequality gap in technological capabilities~\cite{Risks_and_opportunities_2024}. However, as discussed above, equal access to \ac{AI} does not imply equity; there remains a gap in capabilities, knowledge and human resources between different entities, social groups, and geographical areas that opening access does not address~\cite{Open-Sourcing_Highly_Capable_Models_2023,open_for_business_2023}.

\subsubsection{Diversity}
\label{sec:diversity}

While removing discriminatory barriers generally improves diversity, it does not guarantee it. For some concepts of openness, diversity must be actively sought and encouraged. For example, \textit{openness to experience} describes individuals who are not only tolerant of differences but also actively seek out novelty and diversity~\cite{openness_to_experience_3}. Similarly, the variety of colloquiums offered by \textit{open universities} permits to attract diverse profiles of students~\cite{open_university_2_2016}. One of the benefits of diversity is the multiplicity of perspectives, which is particularly valued in \textit{open strategy} by involving a wide range of stakeholders in the formulation and implementation of corporate strategy~\cite{Open_strategy_1_2019,open_strategy_2}. Diversity also includes representativeness and adaptation to minorities.

\textbf{\textit{Diversity} in \ac{AI} Openness:} The adaptability of open models enables to address a broader diversity of needs, thereby making \ac{AI} accessible to marginalized populations. For instance, large language models do not support or perform well with low-resource languages~\cite{systematic_inequalities_in_language_technology_2021}, but the customization potential of open models is promising for bridging this gap~\cite{typhoon_thai_llm_2023,performance_of_llm_low_resources_languages_2024,Hazards_accessible_fine_tuning_2023}. However, like fairness, diversity in \ac{AI} openness is considered more as a consequence of openness than a foundational principle.

\subsubsection{Democratization}
\label{sec:democratization}

Inclusion not only includes being fair and ensuring diversity, it also refers to democratizing and redistributing power, knowledge, or resources across the population. For example, \textit{open access}~\cite{Open_access_1_2015,Open_access_2_2020}, \textit{massive open online course}~\cite{MOOC_1}, and \textit{urban open space}~\cite{urban_open_space_3,urban_open_space_4} all refer to access that is public, available to all individuals without distinctions, unlike \textit{open innovation}, which can be limited to only a few companies~\cite{open_innovation_3,Open_innovation_2_2018}. But public access is often not enough to democratize or redistribute power, the empowerment and democracy may also be necessary. For example, \textit{open education} refers not only to public education but also to democratic education, emphasizing the importance of redistributing power to individuals in addition to public access~\cite{open_education_3}. Another example is \textit{open science}, which aims to democratize knowledge, empower end users by allowing them to be producers of ideas, and make science truly accessible to everyone~\cite{open_science_2,Open_science_1_2023}.

\textbf{\textit{Democratization} in \ac{AI} Openness:} While some previous studies refer to models as ``open model'' when their weights are ``widely available,'' with some potential exceptions (e.g. age limit)~\cite{on_the_societal_impact_2024}, others define an open model as publicly available, with no restrictions on who can access to the components~\cite{Open-Sourcing_Highly_Capable_Models_2023,OSI_Open_Source_AI_def}. However, public access does not guarantee democratization or the redistribution of power. The group of stakeholders most empowered by openness in AI are developers, particularly those creating downstream applications with powerful models~\cite{Risks_and_opportunities_2024}. 
But, while some argue that openness reduces monopolies by enabling small companies or groups of developers to reuse models they would not have the resources to develop on their own~\cite{on_the_societal_impact_2024,Open-Sourcing_Highly_Capable_Models_2023,considerations_for_governing_open_2024}, others counter that openness empowers large companies by making downstream applications dependent on their models and increase their competitive advantage~\cite{Why_open_are_actually_close_2024,why_companies_democratise_AI_2024}.

\section{Discussion}

Our taxonomy provides a systematic lens through which openness could be more broadly understood for the AI community. 
We find that openness in AI emphasizes \textit{access}, \textit{inspectability}, \textit{reuse}, \textit{organic} and \textit{democratization} but does not represent some other possible notions of openness, such as \textit{non-isolation}, \textit{autonomy}, \textit{fairness}, and \textit{diversity}, as other concepts do. 
Many of these openness sub-themes are interconnected, just like how \textit{open strategy} -- the involvement of external individuals in strategy development (i.e., collaboration) -- enables the diversification of perspectives (i.e., diversity), the redistribution of power (i.e., democratization), and relies on the disclosure of information that is usually private (i.e., inspection)~\cite{Open_strategy_1_2019}.  
Similar connections between the themes can be made in AI: grouping together a wide range of practices and principles under the concept of AI openness can enable a more holistic discussion that facilitates the discovery of links and relationships between the themes. 

However, some of the themes addressed in the taxonomy may be undesirable when applied to AI. For example, the sub-theme “organic” in our taxonomy, when applied to AI, would refer to a reduction in governance and the removal of potential regulations, including those on the downstream uses of AI models. However, the absence of regulation can lead to irreversible risks, and it is reasonable to seek to prevent these risks through restrictions on possible uses, as exemplified by Responsible AI Licenses~\cite{RAIL_uses_2024,RAIL_intro_2022}. Given the positive connotation of openness supported by many governments and industry stakeholders, it is crucial to shape this notion so that it aligns with the ethical standards. The definition of AI openness can shape regulation, shift legal liability, encourage some AI development practices at the expense of others. The framework used to define and assess AI openness therefore plays a crucial role in preventing irreversible risks, and attenuate or demand certain notions of openness in AI over others.
It can also be used to tangibly shift the term ``openness'' in AI from a marketing ploy to ethical guarantees.

\subsection{Properties and Desired Effect}
\label{sec:properties_effects}

In the \ac{AI} community today, ``open'' and ``open source'' primarily refer to the actions afforded by the \ac{AI} system (e.g. being able to access, use, share, modify, study). 
Both the definition of Open Source AI from \ac{OSI}~\cite{OSI_Open_Source_AI_def} and the Model Openness Framework~\cite{Model_Openness_Framework_2024} are examples of this; they define AI openness in terms of who can perform these actions (e.g. anyone), how they can be performed (e.g. without any fees or the need for permission), and on what part of the AI system (e.g. model weights, inference code, training code). Defining openness solely through \textit{afforded actions} presents two limitations. First, these actions can be unattainable ideals. For example, 
even with detailed documentation and access to all system components, a large portion of the population would still be unable to properly study an AI system, so this \textit{afforded action} is difficult to achieve for everyone. In contrast, clear properties such as ``being accompanied by a model card'' are objectively verifiable and implementable. Therefore, defining openness through the \textit{properties} of the AI system could be valuable in establishing efforts toward common practices, such as implementing standardized interfaces or providing modular components for AI development. 

The second limitation of defining openness through \textit{afforded actions} is that their ethical impacts  might not be achieved. Impacts such as democratization of AI development~\cite{Risks_and_opportunities_2024,why_companies_democratise_AI_2024} and transparency of AI model~\cite{Openwashing_and_UE_Act_2024,Why_open_are_actually_close_2024,on_the_societal_impact_2024} are presented as consequences of openness in previous work but might be at stake depending on the context of how openness is realized. For instance, being able to inspect a model does not always guarantee auditability~\cite{open_for_business_2023}. On the contrary, if openness in AI is defined by the \textit{desired effect}, e.g., being audited, it could create direct incentives for these audits. We therefore recommend using all three approaches when defining openness in AI (\textit{properties}, \textit{afforded actions} and \textit{desired effect}) to clearly specify the actions, the goal of the actions, and the properties that enable those actions. 

\subsection{Openness at Different Scales}
\label{sec:scale}

\textit{Openness} as an overarching notion can be applied to different scales within a single field. For example, in education, \textit{open educational resources}~\cite{open_educational_ressources_1,open_educational_ressources_4}, \textit{open university}~\cite{open_univeristy_1,open_university_2_2016}, and \textit{open education}~\cite{open_education_1,open_education_2} respectively refer to the openness of resources, universities, and the field of education as a whole. Moreover, concepts at different level can interact and contribute to each other, helping to construct a comprehensive understanding of openness. For example, \textit{open education} includes the use of \textit{open educational resources} along with other principles applicable at the scale of the education field. This hierarchy within the concepts allows for compatibility between them, which can then simplify corresponding regulations and policies. It also highlights the missing parts needed to generalize an effect from one scale to another: for example, the accessibility of educational resources cannot be generalized to the accessibility of education, which faces additional challenges such as providing education to students that lack of free time to study, independently of access to educational resources~\cite{open_education_1}.

While discussions in \ac{AI} largely focus on the openness of \ac{AI} models or systems, distinct openness concepts could be used to address the openness at the level of \ac{AI} organizations, \ac{AI} development processes, \ac{AI} governance models, or the \ac{AI} field itself. Topics like democratization~\cite{Democratizing_AI_2023} and participation~\cite{Beyond_participatory_2024} have already been discussed in different contexts. Connecting them with AI openness at the levels beyond the model could facilitate a holistic view of their impact. Similarly to how accessibility of open educational resources does not extend to accessibility of education, it would also be relevant to measure to what extent the openness of AI systems allows the field itself to open up, or whether other properties need to be enforced at the field level to meet such an objective.

\subsection{Limitations}

The methodology we applied to identify openness concepts in the literature through topic modeling does not produce an exhaustive list of concepts related to openness; therefore, some notions of openness may have been overlooked in our taxonomy. This may be the case for openness concepts that are under-represented in the academic literature, concepts with a highly heterogeneous associated vocabulary, or those that are primarily discussed in languages other than English. Despite these limitations, the concepts collected in this study represent a wide range of openness notions -- sufficient for the results of our analysis to provide meaningful insights on openness in \ac{AI}.

\section{Conclusion}

This study contributes to the ongoing discourse on AI openness by broadening its scope through interdisciplinary perspectives. Our taxonomy, developed from a qualitative analysis of diverse openness concepts, highlights the relevance of the transdisciplinary notion of openness to AI beyond traditional open source paradigms. By emphasizing the themes of \textit{interactivity}, \textit{freedom}, and \textit{inclusiveness} and describing different approaches to define openness, our taxonomy provides a framework to re-evaluate principles and practices of openness in AI. We encourage future work to address underexplored themes of AI openness such as \textit{collaboration}, \textit{non-isolation} and \textit{diversity}, and enrich AI openness at all scales, ranging from defining concrete \textit{properties} of open models to discussing \textit{desired effects} of openness at the scale of the AI field.

\begin{acks}
We acknowledge the financial support from McCall MacBain Scholarship program, the Natural Sciences and Engineering Research Council of Canada (NSERC), and Google. We thank Shalaleh Rismani, Michael Noukhovitch, Ian Arawjo, and Antoine Bou Khalil for their feedback and comments that help us improve the quality of this work.
\end{acks}

\bibliographystyle{ACM-Reference-Format}
\bibliography{bibli}

\appendix

\section{Appendix: Details of the Topic Model}

\label{sec:appendix}

Here we provide further details on the evaluation methods, hyperparameters and results of the topic model used to extract the openness concepts. A complete list of topics and associated openness concepts is provided in Section \ref{sec:list_of_topics}.

\subsection{Evaluation}

We computed the stability of each topic to identify topics that failed to appear in other seeds, thereby improving the reproducibility and robustness of our results. The stability score computed is based on the cosine similarity of the terms-topics distances with the topics of 5 other seeds and matching topics using a recurrent approach (one-to-many)~\cite{Review_stability_topic_modeling_2024}. We chose this method to reduce impact on the stability score of topics splitting or merging across seeds and to only detect topics that failed to appear in other seeds.

We also computed several coherence measures based on the work of~\citet{topic_coherence_2015}. The coherence and stability measures allowed us to narrow down the range of hyperparameter values, and the quality of the topics was then manually evaluated at each iteration to select a number of topics relevant to our analysis.

\subsection{Hyperparameters}

\begin{table}
    \caption{Ranges of values tested for each hyperparameter and the corresponding selected values.}
    \label{tab:hyperparameters}
    \begin{tabular}{ccc}
        \toprule
        Hyperparameter&Range tested&Value selected\\
        \midrule
        k (number of topics)&10 to 200&80\\
        alpha (doc-topic prior)&0.01/k to 10/k&0.1/k\\
        eta (topi-word prior)&0.001 to 5&0.1\\
        \bottomrule
    \end{tabular}
\end{table}

Table \ref{tab:hyperparameters} summarizes the range of hyperparameter values that were tested and those that yielded the best results. We assumed that most abstracts contained only a single topic, which is confirmed by the selected alpha value. Similarly, although some terms appear in multiple topics, we found that most topics used a specific vocabulary, which is reflected in the chosen eta value. 

We used a Gibbs-sampling-based LDA and set the number of iterations to 1000 and burn-in of 100 iterations.

\subsection{List of Topics}
\label{sec:list_of_topics}
Tables \ref{tab:topicmodeling1}, \ref{tab:topicmodeling2}, and \ref{tab:topicmodeling3} present, for each topic, the list of the 10 terms closest to the topic (top terms), the stability of the topic, and the openness concepts that were analyzed in each round of qualitative analysis.

Some topics have the same associated openness concepts, so we ignored their second occurrence after verifying that it was not a homonym. In addition, no openness concepts were analyzed for eight topics with very low stability (below 0.63) to retain only the topics most representative of the dataset.

\begin{table*}
  \caption{Topics identified by LDA analysis (1 to 27) and their associated openness concepts used for the reflexive thematic analysis.}
  \label{tab:topicmodeling1}
  \begin{tabular}[x]{cp{7.8cm}cp{3cm}p{3cm}}
        \toprule
        \multicolumn{1}{l}{\textbf{ID}}&
        \multicolumn{1}{c}{\textbf{Top terms}}&
        \multicolumn{1}{c}{\textbf{Stability}}&
        \multicolumn{2}{c}{\textbf{Openness concept analyzed}}\\ 
        
        \cmidrule(lr){4-5}
        &
        &
        &
        \multicolumn{1}{c}{\textbf{Round 1}}&
        \multicolumn{1}{c}{\textbf{Round 2}}\\
        \midrule
        1&fault, voltage, inverter, phase, circuit, propose, switch, current, motor, converter&0.95&Open circuit&Open phase fault\\
        2&nephrectomy, renal, donor, prostatectomy, prostate, biopsy, opn, kidney, partial nephrectomy, radical&0.57&&\\
        3&group, laparoscopic, surgery, operation, postoperative, open surgery, cholecystectomy, time, control group, case&0.85&Open surgery&Open thyroidectomy\\
        4&hernia, group, laparoscopic, repair, mesh, appendectomy, operative, inguinal, hernia repair, patient&0.81&Open appendectomy&Open hemorrhoidectomy\\
        5&source, open source, software, technology, development, project, open, system, design, source software&0.74&Open source (philosophy)&Open standard\\
        6&risk, injury, globe, factor, open globe, risk factor, globe injury, open globe injury, study, visual&0.55&&\\
        7&web, information, datum, service, user, system, link, gis, application, semantic&0.70&Open hypermedia&Open API\\
        8&heat, temperature, air, water, thermal, pressure, wind, cool, fuel, gas&0.74&Open cycle&Open cathode\\
        9&market, policy, economic, price, country, china, law, political, financial, stock&0.70&Open market&Open banking\\
        10&datum, dataset, information, data, analysis, database, provide, use, available, open datum&0.80&Open data&Open Source Intelligence (OSINT)\\
        11&trial, group, intervention, treatment, participant, randomize, control, woman, clinical, care&0.68&Open label study / trial&\\
        12&space, open space, urban, city, green, area, public, office, sound, green open&0.92&(Public / Urban) open space&Green open space\\
        13&auc, dose, formulation, pharmacokinetic, max, concentration, healthy, subject, plasma, sequence&0.82&Open reading frame&\\
        14&string, shell, theory, spin, open shell, state, open string, energy, brane, field&0.67&Open shell&Open string\\
        15&blood, group, heart, level, serum, cpb, surgery, pressure, open heart, heart surgery&0.75&Open heart surgery&\\
        16&frequency, antenna, filter, resonator, ghz, band, waveguide, dielectric, mode, wave&0.88&Open resonator&Open loop resonator\\
        17&trade, openness, trade openness, economy, growth, economic, country, financial, economic growth, policy&0.94&Trade openness&Open economy\\
        18&bite, open bite, anterior, tooth, anterior open, anterior open bite, treatment, implant, skeletal, orthodontic&0.86&Open bite&Open apex\\
        19&image, use, robot, system, sensor, device, signal, measurement, accuracy, detection&0.73&Open source hardware&\\
        20&heart, open heart, surgery, heart surgery, open heart surgery, patient, icu, blood, cardiac, suction&0.79&&\\
        21&student, open end, end, thinking, learn, question, mathematical, ability, test, solve&0.77&Open ended question/task&Open inquiry\\
        22&power, loop, energy, control, open loop, system, voltage, grid, power system, propose&0.78&Open loop (control)&\\
        23&educational, education, oer, social, learn, practice, open, article, work, resource&0.73&Open educational ressources&Open mindness\\
        24&mis, lumbar, tlif, fusion, invasive, minimally invasive, minimally, screw, spinal, spine&0.56&&\\
        25&lung, pulmonary, valve, heart, cardiac, chest, patient, biopsy, ventilation, mitral&0.68&Open lung biopsy&\\
        26&wound, infection, fracture, antibiotic, open fracture, injury, debridement, tissue, trauma, ssi&0.74&Open fracture&Open wound\\
        27&cell, circuit, solar cell, open circuit, solar, voltage, circuit voltage, open circuit voltage, film, device&0.92&&\\
        \bottomrule
    \end{tabular}
\end{table*}

\begin{table*}
  \caption{Topics identified by LDA analysis (28 to 54) and their associated openness concepts used for the reflexive thematic analysis.}
  \label{tab:topicmodeling2}
  \begin{tabular}[x]{cp{8cm}cp{3cm}p{3cm}}
        \toprule
        \multicolumn{1}{l}{\textbf{ID}}&
        \multicolumn{1}{c}{\textbf{Top terms}}&
        \multicolumn{1}{c}{\textbf{Stability}}&
        \multicolumn{2}{c}{\textbf{Openness concept analyzed}}\\ 
        
        \cmidrule(lr){4-5}
        &
        &
        &
        \multicolumn{1}{c}{\textbf{Round 1}}&
        \multicolumn{1}{c}{\textbf{Round 2}}\\
        \midrule
        28&plant, yield, dry, soil, fruit, crop, production, open field, field, concentration&0.80&Open field (agriculture)&Open pollinate\\
        29&protein, cell, orf, gene, channel, bind, dna, sequence, virus, expression&0.86&Open complex&Open chromatin\\
        30&arthroscopic, knee, shoulder, degree, osteotomy, tibial, wedge, medial, repair, cuff&0.74&Open wedge tibial osteotomy&Open latarjet procedures\\
        31&model, algorithm, optimization, propose, time, parameter, cost, optimal, vehicle, prediction&0.76&Open Shop Scheduling Problem&Open Vehicle Routing Problem\\
        32&education, teaching, university, teach, open, student, open education, open university, laboratory, development&0.77&Open education&Open laboratory\\
        33&patient, chemotherapy, survival, plus, progression, cancer, response, treatment, free survival, month&0.85&&\\
        34&treatment, week, symptom, disorder, depression, score, scale, improvement, patient, baseline&0.91&&\\
        35&tool, open source, software, source, model, code, simulation, package, user, use&0.81&Open source (software/library)&\\
        36&patient, month, year, fol, surgery, mean, age, score, hip, outcome&0.52&&\\
        37&rat, open field, activity, animal, brain, test, behavior, field, mouse, effect&0.84&Open field (test / behavior) (neurology)&Open skill\\
        38&pain, group, insulin, analgesia, opioid, glucose, anesthesia, block, control, patient&0.72&&\\
        39&group, patient, day, treatment, center dot, dot, trial, therapy, assign, hiv&0.80&&\\
        40&system, nerk, service, architecture, platform, application, cloud, security, communication, control&0.90&Open architecture (computing)&Open system (computing)\\
        41&software, oss, project, source, open source, developer, source software, open source software, code, development&0.91&Open source (software) development&\\
        42&channel, velocity, open channel, cavity, pile, turbulent, bed, turbulence, number, wall&0.85&Open channel&Open cavity\\
        43&class, set, domain, recognition, model, dataset, task, open set, unknown, propose&0.90&Open set (Machine Learning)&Open domain (Machine Learning)\\
        44&abdominal, closure, technique, surgical, patient, case, procedure, complication, abdomen, catheter&0.77&Open abdomen&Open drainage\\
        45&pit, mine, open pit, mining, slope, rock, coal, pit mine, open pit mine, stability&0.94&Open pit mine&Open(-off) cut\\
        46&cancer, dose, tumor, response, breast, patient, phase, arm, cell, breast cancer&0.92&&\\
        47&meta analysis, meta, vat, cost, outcome, survival, resection, analysis, orc, compare&0.70&Open thoracotomy&Open lobectomy\\
        48&system, quantum, loop, state, dynamic, control, open loop, open system, quantum system, equation&0.77&Open quantum system&\\
        49&fracture, union, infection, tibial, open fracture, fixation, tibia, nail, flap, external&0.87&&\\
        50&specie, forest, area, habitat, water, site, land, tree, vegetation, species&0.85&Open water&Open habitat\\
        51&innovation, open innovation, firm, knowledge, business, company, sme, open, enterprise, external&0.98&Open innovation&Open business model\\
        52&patient, complication, risk, undergo, cost, associate, mortality, cohort, hospital, outcome&0.84&Open radical prostatetomy&\\
        53&cervical, door, laminopy, open door, spinal, group, door laminopy, open door laminopy, calcaneal, foot&0.68&Open-door laminopy&\\
        54&week, patient, ole, seizure, safety, treatment, dose, efficacy, remission, inc&0.83&&\\
        \bottomrule
    \end{tabular}
\end{table*}

\begin{table*}
  \caption{Topics identified by LDA analysis (55 to 80) and their associated openness concepts used for the reflexive thematic analysis.}
  \label{tab:topicmodeling3}
  \begin{tabular}[x]{cp{8cm}cp{3cm}p{3cm}}
        \toprule
        \multicolumn{1}{l}{\textbf{ID}}&
        \multicolumn{1}{c}{\textbf{Top terms}}&
        \multicolumn{1}{c}{\textbf{Stability}}&
        \multicolumn{2}{c}{\textbf{Openness concept analyzed}}\\ 
        
        \cmidrule(lr){4-5}
        &
        &
        &
        \multicolumn{1}{c}{\textbf{Round 1}}&
        \multicolumn{1}{c}{\textbf{Round 2}}\\
        \midrule
        55&flux, pregnancy, fetal, magnetic, birth, woman, maternal, solar, pregnant, bifida&0.30&&\\
        56&innovation, open innovation, knowledge, firm, factor, openness, social, relationship, influence, research&0.85&Open strategy&\\
        57&crack, rotor, hole, strength, load, test, composite, failure, steel, stress&0.71&Open hole&Open rotor\\
        58&health, openness, participant, experience, care, survey, child, self, study, family&0.77&Openness to experience&Open dialogue\\
        59&water, ocean, sea, concentration, emission, sediment, open ocean, ice, carbon, surface&0.76&Open ocean (continent)&Open water / ocean (ice)\\
        60&structure, metal, reaction, compound, column, crystal, capillary, framework, angstrom, adsorption&0.81&Open framework&Open-tubular\\
        61&poag, eye, glaucoma, thickness, visual, correlation, visual field, rnfl, retinal, optic&0.88&Open angle glaucoma&\\
        62&hole, oil, reservoir, open hole, drill, completion, gas, log, drilling, fracture&0.88&Open hole (well)&\\
        63&iop, glaucoma, eye, open angle, angle, angle glaucoma, open angle glaucoma, mmhg, poag, intraocular&0.87&&\\
        64&foam, cell, open cell, section, pore, material, beam, porosity, structure, property&0.76&Open cell foam&Open (cross) section\\
        65&repair, aortic, aneurysm, endovascular, evar, mortality, aaa, aortic aneurysm, artery, patient&0.90&Open repair&\\
        66&journal, access, open access, article, publish, publication, citation, review, author, research&0.92&Open access&Open peer review\\
        67&laser, sample, gas, measurement, column, liquid, collision, plasma, charm, ion&0.59&&\\
        68&patient, week, treatment, dose, safety, adverse, efficacy, adverse event, study, event&0.88&&\\
        69&iop, glaucoma, ocular, eye, pressure, timolol, mmhg, latanoprost, open angle, intraocular pressure&0.87&&\\
        70&datum, government, research, science, open datum, public, open science, open, policy, ogd&0.78&Open science&Open government\\
        71&covid, vaccine, vaccination, dose, cov, sar, sar cov, antibody, pandemic, participant&0.60&&\\
        72&student, learn, course, online, education, learner, learning, mooc, university, oer&0.95&Massive Open Online Couse (MOOC)&Open university\\
        73&boundary, equation, wave, field, condition, solution, boundary condition, numerical, model, open boundary&0.73&Open boundary&Open surface\\
        74&laparoscopic, resection, group, surgery, cancer, gastrectomy, gastric, survival, postoperative, patient&0.92&&\\
        75&fracture, reduction, carpal, tendon, fixation, tunnel, carpal tunnel, release, open reduction, nerve&0.63&&\\
        76&fracture, fixation, reduction, open reduction, orif, internal fixation, internal, case, reduction internal, reduction internal fixation&0.93&Open reduction&\\
        77&poag, glaucoma, gene, mutation, genetic, polymorphism, open angle, variant, open angle glaucoma, angle glaucoma&0.85&&\\
        78&cluster, star, open cluster, ngc, mass, stellar, binary, age, galactic, member&0.97&Open cluster&\\
        79&access, open access, library, journal, publish, university, payment, author, repository, article&0.90&Open payment&\\
        80&set, space, open set, fuzzy, topological, soft, open, function, property, graph&0.86&Open set (mathematics)&Open book (mathematics)\\
        \bottomrule
    \end{tabular}
\end{table*}

\end{document}